\documentclass[10pt,twocolumn,letterpaper]{article}

\usepackage{ijcb}
\usepackage{times}
\usepackage{epsfig}
\usepackage{graphicx}
\usepackage{amsmath}
\usepackage{amssymb}
\usepackage{enumitem}
\usepackage{xcolor}
\usepackage{float}
\restylefloat{table}

\usepackage[pagebackref=true,breaklinks=true,letterpaper=true,colorlinks,bookmarks=false]{hyperref}
\usepackage[ruled, lined, linesnumbered, commentsnumbered, longend]{algorithm2e}

\ijcbfinalcopy 

\ifijcbfinal\fi
\begin{document}

\title{Federated Learning-based Active Authentication on Mobile Devices}

\author{Poojan Oza \hspace{6mm} Vishal M. Patel\\
	Department of Electrical and Computer Engineering\\
	Johns Hopkins University, 3400 N. Charles St., Baltimore, MD 21218, USA\\
	{\tt\small $\lbrace$poza2, vpatel36$\rbrace$@jhu.edu}
}

\maketitle
\thispagestyle{empty}

\begin{abstract}
User active authentication on mobile devices aims to learn a model that can correctly recognize the enrolled user based on device sensor information. Due to lack of negative class data, it is often modeled as a one-class classification problem. In practice, mobile devices are connected to a central server, e.g, all android-based devices are connected to Google server through internet. This device-server structure can be exploited by recently proposed Federated Learning (FL) and Split Learning (SL) frameworks to perform collaborative learning over the data distributed among multiple devices. Using FL/SL frameworks, we can alleviate the lack of negative data problem by training a user authentication model over multiple user data distributed across devices. To this end, we propose a novel user active authentication training, termed as \textit{Federated Active Authentication (FAA)}, that utilizes the principles of FL/SL. We first show that existing FL/SL methods are suboptimal for FAA as they rely on the data to be distributed homogeneously (i.e. IID) across devices, which is not true in the case of FAA. Subsequently, we propose a novel method that is able to tackle heterogeneous/non-IID distribution of data in FAA. Specifically, we first extract feature statistics such as mean and variance corresponding to data from each user which are later combined in a central server to learn a multi-class classifier and sent back to the individual devices. We conduct extensive experiments using three active authentication benchmark datasets (MOBIO, UMDAA-01, UMDAA-02) and show that such approach performs better than state-of-the-art one-class based FAA methods and is also able to outperform traditional FL/SL methods.
\end{abstract}

\section{Introduction}\label{sec:introduction}

\begin{figure}[t!]
	\centering
	\includegraphics[width=0.50\linewidth]{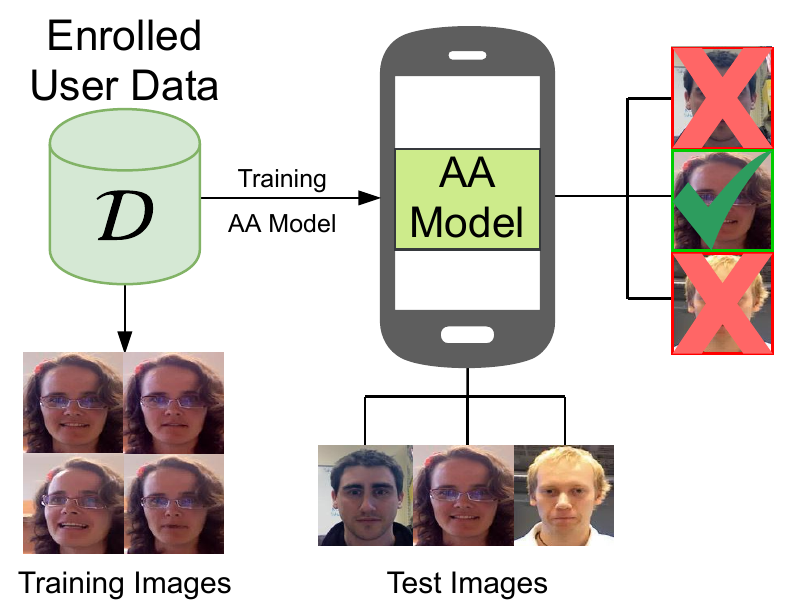}
	\vskip -6.0pt \caption{An overview of face-based active authentication.}
	\label{fig:intro_figure}
\end{figure}

Recent years have seen a surge in the number of mobile device users in the world. According to an estimate, nearly 60\% of the world's population own at least one mobile device\footnote[1]{www.bankmycell.com/blog/how-many-phones-are-in-the-world}. This increasing popularity is driven by the flexibility and convenience of mobile devices in handling many important tasks such as banking, finance management, social media, navigation etc. The handing of these tasks require users to store their personal information such as bank account details, social media profiles and other passwords in the mobile device. As a result, when mobile devices are lost or stolen it can compromise sensitive and private information of the user. Hence, increasing the security of mobile devices and protecting private information of users becomes extremely important. Most of the conventional authentication approaches prompt the user initially and grant access to the device until the session is over (i.e. explicit authentication \cite{patel2016continuous}). As long as the mobile phone remains active, typical devices incorporate no mechanisms to verify that the user originally authenticated is still the user in control of the device. Thus, unauthorized individuals may improperly obtain access to a user's personal information if a password is compromised or if the user does not exercise adequate vigilance after initial authentication.

\begin{figure*}[t!]
	\centering
	\includegraphics[width=0.70\linewidth]{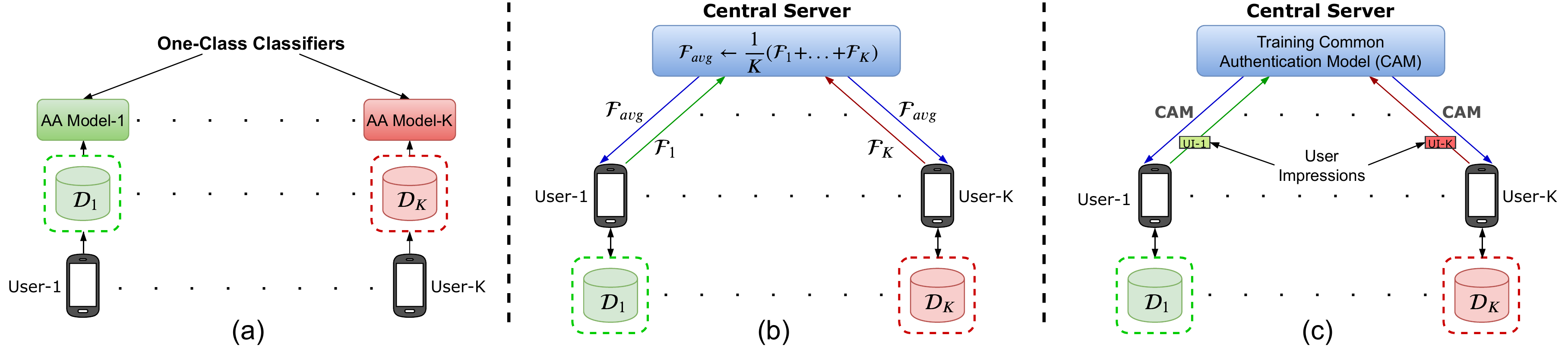}
	\vskip -6.0pt \caption{ Active authentication based on (a) One class classification, (b) Federated averaging, and (c) Proposed method.}
	\label{fig:oc_vs_fl}
\end{figure*}

User active authentication was specifically introduced to address this issue \cite{patel2016continuous}. In user active authentication, built-in sensor information such as touch-sensor, gyroscope, microphone, accelerometer, camera, finger-print reader  etc. are used to continuously monitor user \cite{patel2016continuous}. These sensors can capture behavioral (touch patterns, accelerometer etc.) or physiological (fingerprint, face etc.) biometric traits of the user. In this paper, we focus on developing active authentication systems based on face images collected by the front-facing camera of the device.

One way to develop such an authentication system would be to combine all user data from different devices in a central server and develop a multi-class classification model. However, it is not feasible to transfer mobile device data from multiple users to a server due to user data-privacy concerns. Most work in the literature overcome this issue by modeling active authentication as a one-class classification problem \cite{perera2018dual}, \cite{perera2017extreme}, \cite{scholkopf2001estimating}, \cite{tax2004support}. As illustrated in Fig.~\ref{fig:oc_vs_fl}(a), in one-class based authentication systems only the data from an enrolled user is used to train the classifier which is later used to detect unauthorized users. However, such models often rely on prior assumptions and can be suboptimal due to the absence of negative class data during training.

Federated learning (FL) \cite{mcmahan2016communication} and Split Learning (SL) \cite{gupta2018distributed} frameworks were recently introduced in which mobile phones collaboratively learn a shared authentication model while keeping all the training data private on the individual devices. In the FL/SL framework, the model parameters are shared between server and mobile devices and the user data are kept on the individual devices. This makes information sharing between server-device feasible and addresses the user privacy concerns. The most popular algorithm to train deep neural networks in these frameworks is Federated Averaging (FedAvg) \cite{mcmahan2016communication}. As illustrated in Fig.~\ref{fig:oc_vs_fl}(b), federated averaging algorithm trains a local model for all devices, which are averaged in a central server and sent back to the individual devices. However, the performance of FedAvg heavily relies on the assumption that data distribution across devices is independent and identically distributed, i.e. IID. For example, in a federated learning task of identifying $K$-users, each device is required to contain sufficient data from all $K$ users. When the device data distribution deviates from this IID assumption, (i.e., becomes non-IID), the performance of FedAvg algorithm drops significantly (see \ref{subsec:challenges}). This creates a problem for active authentication where the device data are distributed in a highly non-IID manner. Specifically, in the active authentication setting, each device contains data from only a single user. Hence, directly applying existing FL/SL algorithms will not be useful for federated active authentication.

We propose a novel method for active authentication in the FL/SL framework that addresses these issues by tackling the non-IID nature of the distribution of data among mobile devices. In the proposed method, we first train the full model on an exiting publicly available face recognition dataset. This pre-trained model is then split into a feature extractor network and a classifier network. We share the feature extractor network to all devices which is used to get the feature statistics from data samples of the enrolled user. This makes information sharing between device-server more efficient as we need to share only feature statistics instead of features from the entire dataset. Each device estimates the feature mean and variance, which we call \textit{user impressions}. We send these \textit{user impressions} of all users to a central server where the classifier network is trained on the dataset created by sampling from individual user statistics. This process is illustrated in Fig.~\ref{fig:oc_vs_fl}. We evaluate the proposed method on three  active authentication datasets -- MOBIO \cite{gunther20132013}, UMDAA-01 \cite{fathy2015face} and UMDAA-02 \cite{mahbub2016active}. Experiments show that the proposed method is able perform better than the previous one-class classification-based methods and existing FL/SL approaches. 
	
	This paper makes the following contributions:
\begin{itemize}[topsep=0pt,noitemsep,leftmargin=*]
\item We propose a novel method, termed as \textit{federated active authentication (FAA)}, that utilizes the principles of FL/SL frameworks  to improve user active authentication in a privacy preserving way.
\item We demonstrate the limitations of the existing algorithms and show that the proposed method is able to overcome them for federated active authentication.
\item Extensive experimental analysis on three publicly available datasets (MOBIO, UMDAA-01 and UMDAA-02) show that the proposed method is able to outperform many existing active authentication methods.
\end{itemize}


\section{Related work}\label{sec:related_work}

\subsection{Active authentication}\label{subsec:active_authentication}

Single user active authentication problem has been approached as a one-class classification problem in the literature \cite{perera2018dual}, \cite{perera2017extreme}, \cite{oza2019active}. Most conventional approaches utilize off-the-shelf one-class classification models such as one-class support vector machine (OC-SVM) \cite{scholkopf2001estimating}, support vector data descriptor (SVDD) \cite{tax2004support}, mini-max probability machine (MPM) \cite{lanckriet2002minimax} etc. These one-class classifiers are trained on either hand-crafted features or features extracted from a pre-trained deep neural network. Few recent works attempt to extend these basic one-class classifier formulations by adding more constraints to their objective functions. Noticeably, the work by Perera and Patel  \cite{perera2018dual} extends single mini-max probability machine (SMPM) formulation \cite{ghaoui2003robust} with additional hyperplane constraint to propose a better one-class classifier called dual-minimax probability machines (DMPM). Oza \emph{et al.} \cite{oza2019active} propose to use pseudo-negative data during training with the help of an auto-encoder network architecture to learn a deep neural network based one-class classifier. Many works have explored the use of different biometric modalities such as touch patterns, keystrokes, voice, face for user authentication \cite{dey2016extreme}, \cite{kumar2016continuous}, \cite{frank2012touchalytics}, \cite{serwadda2013verifiers}. More recent works have focused on face-based authentication systems \cite{perera2018dual}, \cite{perera2017extreme},  \cite{oza2019active}.

\subsection{Distributed learning of deep networks}\label{subsec:distributed_learning}
Traditional deep network training assumes that all training data are available at a single data center location for training. In practice that is rarely the case and data are most likely to be distributed among the multiple data centers. Furthermore, these data centers may not allow a direct sharing of their data due to privacy concerns. Federated learning \cite{mcmahan2016communication} and Split Learning \cite{gupta2018distributed} frameworks were specifically proposed to address these issues.

\noindent \textbf{Federated Learning.} Federated learning enables such decentralized deep network training by effectively combining models trained by the individual data centers in a central server connected to all data centers \cite{mcmahan2016communication}. Additionally, such decentralized training protects the privacy of data at individual data centers. This enables a safe collaboration among the data centers to learn a better deep network model without sacrificing user privacy. Federated Averaging (FedAvg) is one of the most widely used  federated learning algorithm to train deep network models \cite{mcmahan2016communication}, \cite{mohri2019agnostic}. In FedAvg, a model is initialized at a central server and sent to all data centers, which then train their individual models with their locally available data. These local models are then sent back to the central server, where all local models' parameters are averaged to create a global model. This global model is then again sent back to the individual data centers for another round of local training and the process is repeated until the global model converges.

\noindent \textbf{Split Learning.} Similar to FL, split learning enables training of deep network when data is shared across multiple devices. Gupta \etal \cite{gupta2018distributed} first introduced split learning where, the whole deep network model is divided into two parts. The first part remains on the local device and the second part is kept on the server. The whole model is trained through backpropagation by passing gradient information from server to local devices. Vepakomma \etal \cite{vepakomma2018split} and Poirot \etal \cite{poirot2019split} utilized split learning framework to train a deep model network on patient data from multiple institutions without having to share the raw patient data. Additionally, Thapa \etal \cite{thapa2020splitfed} proposed an approach that utilizes the principles of both FL and SL to create a fusion method for distributed learning.

However, all of these approaches assume that data is distributed among multiple devices in an IID manner, which does not hold true in the case of FAA. The proposed approach tries to solve this issue by designing an algorithm that can counter the non-IID nature of FAA. Similar to SL \cite{gupta2018distributed}, the proposed approach divides the network into two parts, i.e., feature extractor and classifier. However, the proposed method is more communication efficient as it requires only one round of forward and backward information passing between server and mobile devices as opposed to SL \cite{gupta2018distributed}, which requires multiple rounds. Additionally, instead of directly passing the features of individual data points in mobile devices as done in SL \cite{gupta2018distributed}, we only share feature statistics of the individual mobile device data.

\section{Methodology}\label{sec:methodology}

\subsection{Overview}\label{subsec:overview}
We illustrate the challenges in detail by considering a case study with FedAvg algorithm to show how the performance changes when the IID assumption of the FL/SL framework does not hold. Subsequently, we discuss the proposed solution FAA which overcomes these challenges to provide an improved user authentication system.

\subsection{Challenges}\label{subsec:challenges}

\begin{figure}[t!]
	\centering
	\includegraphics[width=.60\linewidth]{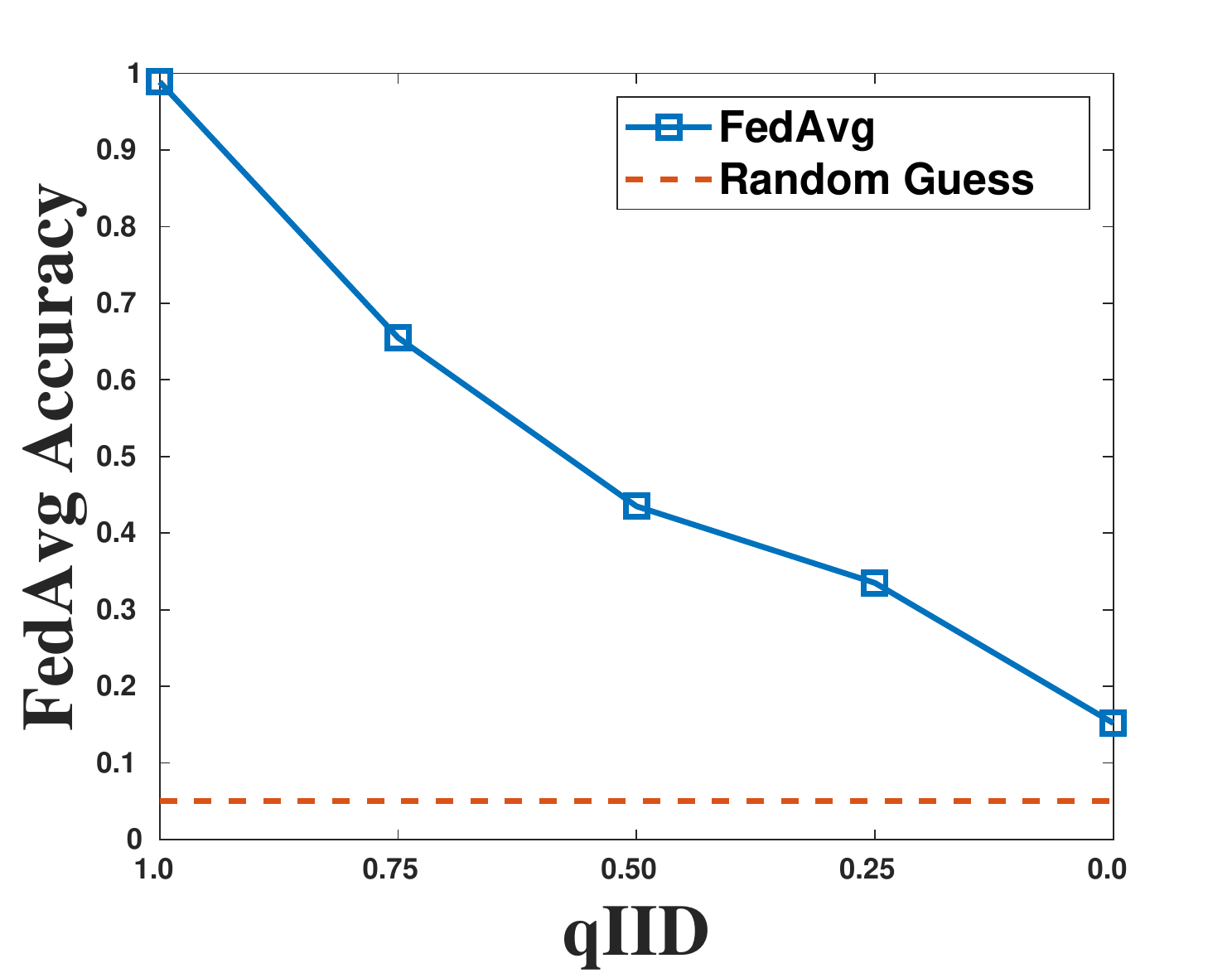}
	\vskip-3.0pt\caption{Performance of federated averaging (FedAvg) algorithm with varying value of $qIID$ representing the way data is distributed among devices.}
	\label{fig:fedavg_weakness}
\end{figure}

\begin{figure*}[t!]
	\centering
	\includegraphics[width=.5\linewidth]{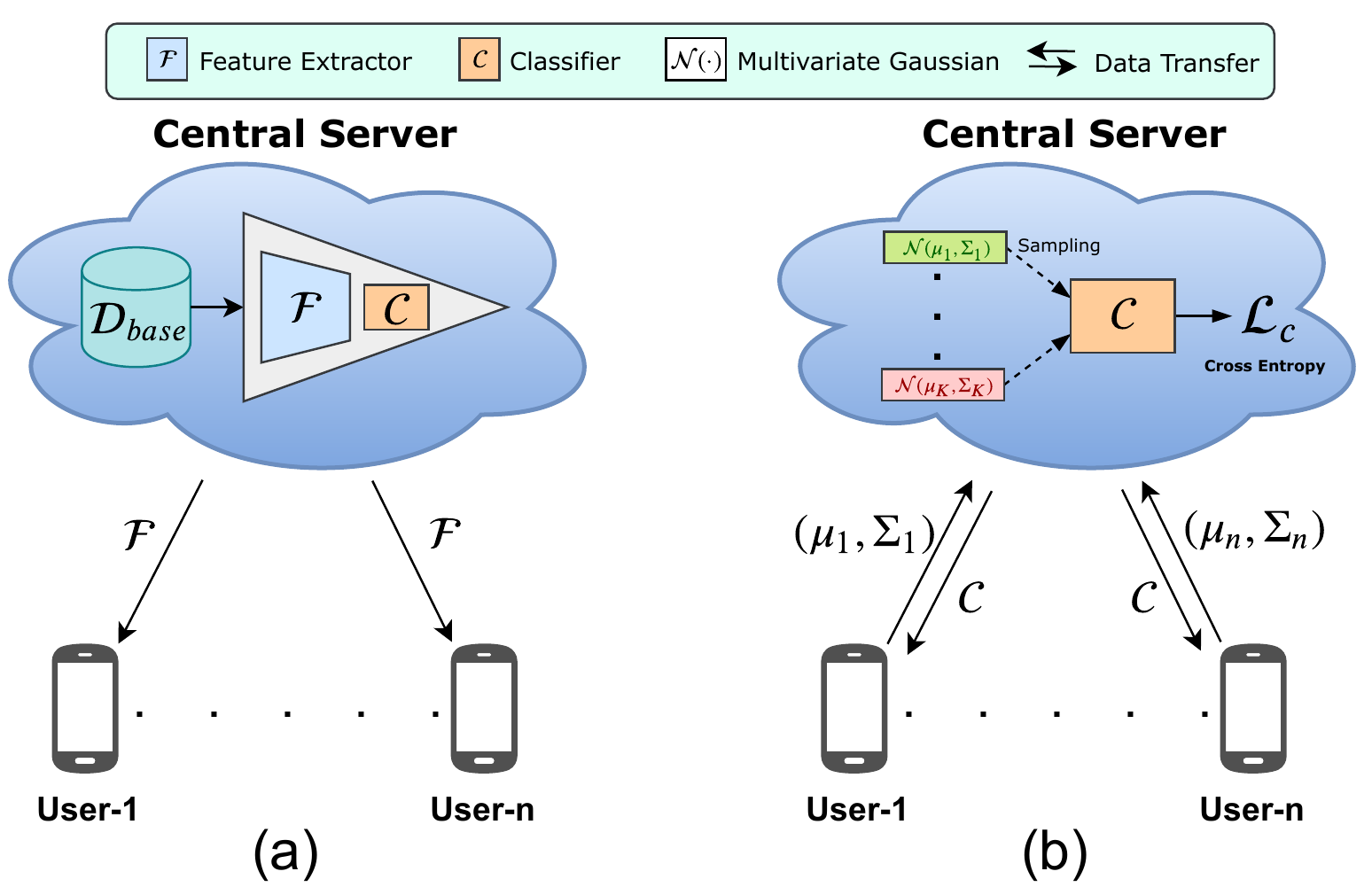}
	\vskip-2.5pt\caption{Block diagram describing the training of the proposed method for federated active authentication. (a) Step-1 of the proposed method: training a model on base dataset, (b) Step-2 \& Step-3 of proposed method: where the we local mobile devices compute user feature statistics and central server trains a classifier, which is sent to the individual devices and used as a authentication model along with the feature extractor.}
	\label{fig:proposed}
\end{figure*}

As explained in the previous section, the key challenge in FAA is the non-IID nature of the data distribution across mobile devices. This issue directly affects the user authentication performance. First, let us briefly discuss the definition of independent and identically distributed data. In the context of federated learning, when data is said to be distributed in an IID manner, it means that each device  has equal number of data samples from all users. This is the most common assumption in federated learning and is very crucial to train a model using the FedAvg algorithm. To show how deviation from this assumption affects the performance of FedAvg algorithm, let us quantify the \textit{IID-ness} of the data distributed among devices. Let us assume that there are $N$ devices containing data from  $K$ users. Let $K_i$ be the number of users contained in the $i^{th}$ device dataset having sufficient number of data samples. Let $qIID$ denote the quantification of ``IID-ness'' of the distributed data in the federated framework. For simplicity, let us assume that each device has equal number of data samples. Given these assumptions, the $qIID$ can be formally written as:
\setlength{\belowdisplayskip}{1pt} \setlength{\belowdisplayshortskip}{1pt}
\setlength{\abovedisplayskip}{1pt} \setlength{\abovedisplayshortskip}{1pt}
\begin{equation}
qIID \ \ = \ \ \frac{\frac{1}{N}\sum_{i=1}^{N} \frac{K_i}{K} - \frac{1}{K}}{1-\frac{1}{K}},
\end{equation}
where $qIID=1$ when the data distribution across devices is the most IID and it decreases as the distribution deviates from  IID. The value  $qIID=0$ represents the most non-IID data distribution across devices. The proposed FAA problem operates on a specific value of $qIID=0$, where the number of devices are equal to the number of users, i.e., $N=K$. To show how the performance of FedAvg algorithm changes when the IID assumption is violated, we perform identification experiments using the UMDAA-01 dataset by changing the $qIID$ value from one to zero. As evident from  Fig.~\ref{fig:fedavg_weakness}, the FedAvg performance heavily relies on the IID assumption. The more distribution of data among devices in the federated learning framework deviates from the IID assumption, the performance of FedAvg degrades significantly. The reason for this reduction in performance is due to averaging of weights at the central server.  This makes sense as the individual models are trained on the data with similar data distributions. Interestingly, for the case of federated active authentication, where the distribution of data among devices is the most non-IID, i.e., $qIID=0$, the performance is almost close to random guessing baseline.

\subsection{Proposed method}\label{subsec:proposed_method}

\subsubsection{Training}\label{subsubsec:training}
\noindent \textbf{Step-1.} Let us first consider  a randomly initialized deep network model $\mathcal{M}$ at the central server. Furthermore, let us denote a publicly available face recognition dataset as, $\mathcal{D}_{base}=\{x_i^{base}, y_i^{base}\}_{i=1}^{N_{base}}$. Here, $x_i^{base}$ are the face images having corresponding labels $y_i^{base}$ where the dataset contains a total of $N_{base}$ images. Note that, $\mathcal{D}_{base}$ does not have any category that overlap with data available in the individual mobile devices. As shown in Fig.~\ref{fig:proposed}(a), the deep network model $\mathcal{M}$ is then trained at server side on the dataset $\mathcal{D}_{base}$ with the help of the following loss:
\begin{equation}
\begin{aligned}
\mathcal{L}_{base} \ = \ \frac{1}{N_{base}} \sum_{i=1}^{N_{base}} \ \mathcal{L}_{c} (\mathcal{M}(x^{base}_i), \ y^{base}_i), 
\end{aligned}
\label{eq:base_loss}
\end{equation}
\noindent where, $\mathcal{L}_{c}$ is the cross-entropy loss function. Once the model $\mathcal{M}$ is trained, it is further divided into two networks, namely, feature extractor network ($\mathcal{F}$) and classifier network ($\mathcal{C}$). The central server  sends feature extractor network $\mathcal{F}$ to all the mobile devices connected to the central server.

\noindent \textbf{Step-2.} Assume that there are $K$ mobile devices (i.e. $K$ users) and the $i$th device has the corresponding dataset $\mathcal{D}_i$ containing $n_i$ face images of the user. All $K$ devices are connected to the central server. With the help of network $\mathcal{F}$, each device estimates the feature mean and variance of the corresponding user, which we refer to as \textit{user impressions}. For the $i$th device, the user impressions can be estimated as:
\begin{equation}
\begin{aligned}
\mu_i \ &= \ \frac{1}{n_i} \sum_{x_j \in \mathcal{D}_i} \mathcal{F}(x_j),\\
\Sigma_i \ &= \ \frac{1}{n_i} \sum_{x_j \in \mathcal{D}_i} (\mathcal{F}(x_j) - \mu_i)(\mathcal{F}(x_j) - \mu_i)^T, 
\end{aligned}
\label{eq:user_impressions}
\end{equation}

\begin{figure*}[t!]
	\centering
	\includegraphics[width=.75\linewidth]{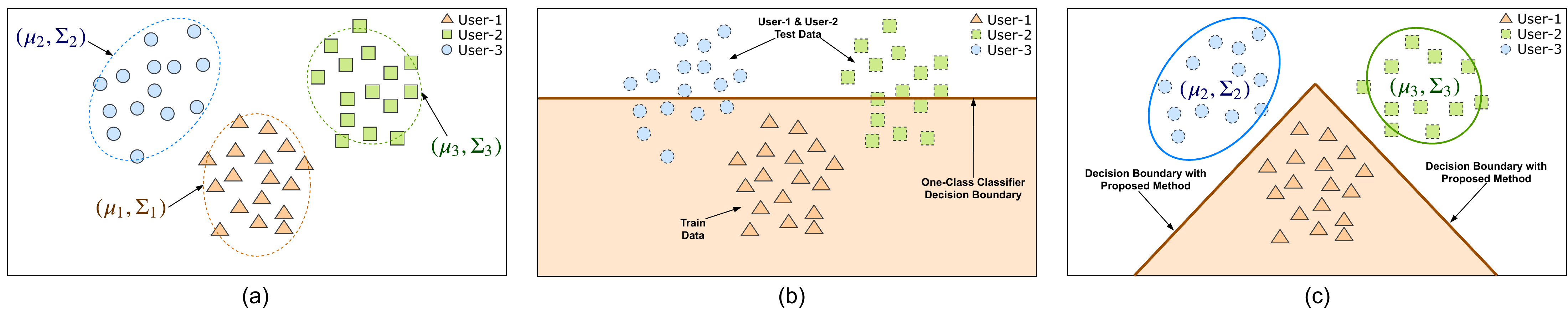}
	\vskip -5.0pt \caption{Toy example with three users to show the effectiveness of the proposed method compared to one-class modeling based methods. (a) Feature space location (mean $\mu_i$) and shape (variance $\Sigma_i$) estimated for each user. (b) Modeling as a one-class classification problem to learn a decision boundary for user-1. When such a model is tested there are many samples from user-2 and user-3 that are mis-classified as user-1. (c) Learning a decision boundary using the proposed method to train the authentication model for user-1 using user-1, user-2 and user-3's mean and variance. This model does not make the same mistake of mis-classifying user-2 and user-3 data as user-1 similar to one-class based method. As can be seen from the figure, the learned decision boundary is also better in comparison to one-class method.}
	\label{fig:toy_example}
\end{figure*}

\begin{algorithm}[b!]\label{algo:training}
	\textbf{Input:} $K, M, E, \eta, \{\mathcal{D}_1,..,\mathcal{D}_N\}, \mathcal{F}, \mathcal{C}$
	
	\While{$i \leq K$}{
		$n_i$ = number of data-points $x_i \in \mathcal{D}_i$,
		
		\While{$k \leq n_i$}{
			Extract features using network $\mathcal{F}$,
			
			$f_k=\mathcal{F}(x_i)$
		}
		
		Extract user impressions for $i^{th}$ user as,
		
		$\mu_i=\frac{1}{n_i}\sum_{k=1}^{n_i} f_k$
		
		$\Sigma_i=\frac{1}{n_i}\sum_{k=1}^{n_i} (f_k-\mu_k)(f_k-\mu_k)^T$
		
		Send $(\mu_i, \Sigma_i)$ of user-$i$ to the central server,
	}
	
	Create approximate feature set for each user-$i$ by sampling $M$ data points from $\mathcal{N}(\mu_i, \Sigma_i)$ ,
	
	$\mathcal{D} = \{f_j \sim \mathcal{N}(\mu_i, \Sigma_i), y_j=i\}$
	
	\While{$e \leq E$}{
		
		Sample batch of features $F_e$ \& label $Y_e$ from $\mathcal{D}$,
		
		Calculate Loss $\mathcal{L}$ as,
		
		$\mathcal{L} = \mathcal{L}_{c}(\mathcal{C}(F_e), Y_e)$
		
		Update the parameters by gradient descent,
		
		$\mathcal{C} \leftarrow \mathcal{C} - \eta*\Delta \mathcal{L}$,
	}
	
	\textbf{Result:} Authentication Model $\{\mathcal{F}, \mathcal{C}\}$
	
	\caption{Algorithm for FAA Training}
	
\end{algorithm}

\noindent where $x_{j}$ is the $j$th face image in $\mathcal{D}_{i}$. Each user impression $(\mu_i, \Sigma_i)$, provides a reasonable estimate regarding the location and the shape of the $i$th user distribution in the feature space of network $\mathcal{F}$. Once all devices have finished estimating user impressions, they are sent to the central server, which creates a Gaussian approximated feature space model of each user as $\mathcal{N}(\mu_i, \Sigma_i)$. This approximation is inspired by the work of Seddik \emph{et al.} \cite{seddik2020random}, which showed that the feature space of deep networks can be well approximated with only first and second order statistics of the features.

\noindent \textbf{Step-3.} With the help of Gaussian approximated feature space models of all users, we create a combined dataset as, $\mathcal{D} = \{f_j \sim \mathcal{N}(\mu_i, \Sigma_i), y_j=i\}$. We make sure that each user has exactly $M$ number of samples, resulting in total $K\times M$ samples. As shown in the Fig.~\ref{fig:proposed}(b), we fine-tune the identification network using the loss given as:
\begin{equation}
\begin{aligned}
\mathcal{L} \ = \ \frac{1}{K\times M} \sum_{j=1}^{K\times M} \ \mathcal{L}_{c} (\mathcal{C}(f_j), \ y_j), 
\end{aligned}
\label{eq:classifier_loss}
\end{equation}
where, $y_j$ is the corresponding user id of feature $f_j$ and $\mathcal{L}_{c}$ is the cross-entropy loss. Once the classifier $\mathcal{C}$ is trained, it is sent to all mobile devices. Both $\mathcal{F}$ and $\mathcal{C}$ together form the authentication system. The full training pseudo-code is described in Algorithm~\ref{algo:training}. Furthermore, in Fig.~\ref{fig:toy_example} we illustrate how the proposed approach is able to utilize user-impressions to improve the authentication model with the help of a toy example with three users. The current state-of-the-art algorithms model the active authentication problem as one-class classification. Due to this, the classifier learned to separate a particular user data still has some risk of failing to restrict the device access to other users, as illustrated in Fig.~\ref{fig:toy_example}(b). However, as shown in Fig.~\ref{fig:toy_example}(c), the proposed approach is able to utilize user impressions from other users to learn a more compact decision boundary and improve the authentication performance.

\subsubsection{Testing}\label{subsubsec:testing}

For any test face image $x_j$, we compute the authentication score corresponding to the $i$th user as,
\begin{equation}
\begin{aligned}
S_j^i \ = \ \mathbb{I}_{[\tilde{y_j} = i]} \ \mathcal{H} [ \mathcal{C}(\mathcal{F}(x_j)) ] \ + \ \mathbb{I}_{[\tilde{y_j} \neq i]} \ \mathcal{H}[q], 
\end{aligned}
\label{eq:test_score}
\end{equation}
where, $\tilde{y}_j$ is the predicted label of the test image $x_j$, i.e., $\text{argmax} \ \mathcal{C} (\mathcal{F}(x_j)) $. The $\mathbb{I}_{[c]}$ is an indicator function which is $1$ when condition $c$ is satisfied and $0$ otherwise. Vector $\mathcal{C} (\mathcal{F}(x_j))$ is a $K \times 1$ prediction vector. The function $\mathcal{H}[\cdot]$ calculates the entropy of the input probability vector. The vector $q$ is $K \times 1$ probability vector with $q_1=q_2=...=q_K=\frac{1}{K}$. When the predicted-id from the authentication model matches the user-id, the first term assigns the score $S_j^i$ as the entropy of the prediction vector, i.e., $\mathcal{C}(\mathcal{F}(x_j))$. When the predicted-id does not match the user-id, the second term penalizes the input for this mis-classification by assigning high entropy value to the score $S_j^i$. When both the terms are added together they encode the score of an input image belonging to the authorized user. Higher score indicates potentially unauthorized user and vice versa.


\section{Experiments and results}\label{sec:experiments_and_results}

\subsection{Implementation details}\label{subsec:implementation_details}

For all experiments, we utilize the VGG16 \cite{simonyan2014very} trained on the VGGFace dataset \cite{Parkhi15}. We consider all conv blocks of VGG16 as the feature extractor network $\mathcal{F}$ and all fully-connected layers as the classifier network $\mathcal{C}$. The mean and variance for each user are estimated by flattening the output of the network $\mathcal{F}$, later used in the server to fine-tune the network $\mathcal{C}$. For training, we utilize SGD optimizer with learning rate 0.001 and momentum 0.9. We train till 100 epochs with the batch size of 64. For all methods, the hyper-parameters are selected based on a validation set. The performance of all methods is evaluated using the average detection accuracy (ADA), defined as:
$$ADA=0.5*(TPR+TNR),$$ where, $TNR$ and $TPR$ represent true negative rate and true positive rate, respectively. 

\subsection{Datasets}\label{subsec:datasets}

\begin{figure}[t!]
	\centering
	\includegraphics[width=\linewidth]{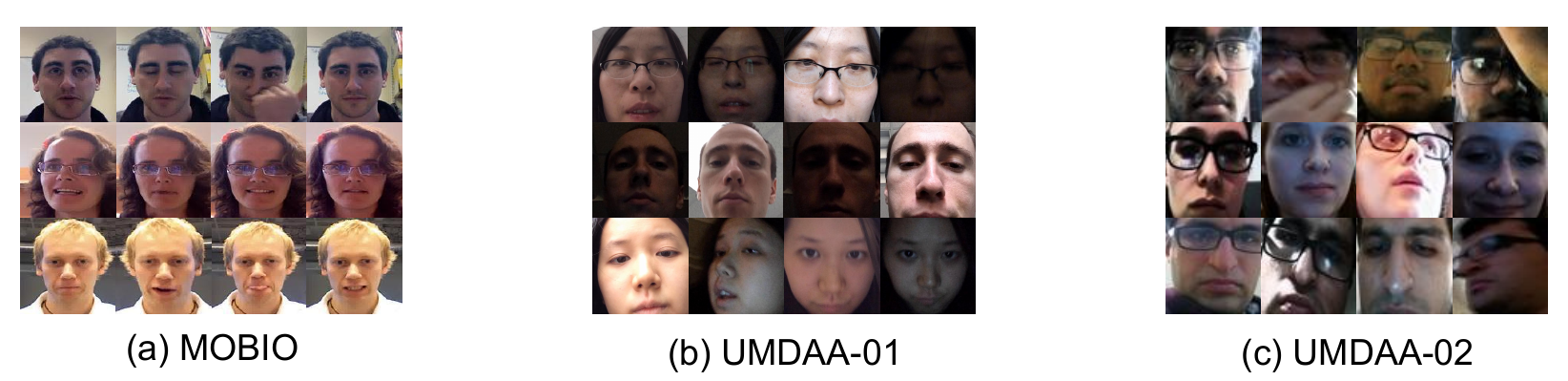}
	\caption{Sample face images from the (a) MOBIO, (b) UMDAA-01 and (c) UMDAA-02 datasets.}
	\label{fig:dataset}
\end{figure}

\noindent \textbf{MOBIO.} The MOBIO \cite{gunther20132013} dataset contains face images and voice data from 150 individuals collected in six different sessions at six different locations. The data is collected using smart phones and/or laptop. For experiments, we only consider the face data. Out of the three datasets, MOBIO is relatively easy as it contains only front facing face images captured in well-lit conditions. Sample images are shown in Fig.~\ref{fig:dataset}(a). The figures provide a reasonable illustration of the variations present in the dataset. For the experiment, we consider the first 75 individuals as the enrolled users and the remaining 75 individuals as unknown/unauthorized users. We create a 50/50 split of data for training and testing for all 150 individuals.\\

\noindent \textbf{UMDAA-01.} The UMDAA-01 \cite{fathy2015face} contains face images of 50 different individuals collected using iPhone 5s in three different sessions with varying lighting conditions. Apart from varying illumination conditions, the dataset also contains multiple other variability in the form of pose, occlusion, facial expressions etc. Sample images are shown in Fig.~\ref{fig:dataset}(b). We consider the first 25 individuals as the enrolled users and the remaining 25 users as unknown/unauthorized. Similar to MOBIO, we create a 50/50 train-test split and use the train split for training.\\

\noindent \textbf{UMDAA-02.} The UMDAA-02 dataset \cite{mahbub2016active} contains information from 18 different sensors such as keystrokes, touch pattern, face images, accelerometer readings from 44 individuals collected using Nexus5 across two months. For this experiment, we only utilize face images of all users. As can be seen from Fig.~\ref{fig:dataset}(c), out of all three datasets, UMDAA-02 contains the most variability in the data samples, proving it to be the most challenging dataset. We consider the first 22 individuals as the enrolled users and the remaining users as unknown/unauthorized.\\

\subsection{Experiments}\label{subsec:experiments}

\begin{table*}[t!]
	\centering
	\huge
	\caption{Performance comparison with state-of-the-art active authentication methods evaluated in terms of average detection accuracy. The best performing method for each dataset is shown in bold fonts.}
	\label{table:ada}
	\resizebox{0.70\linewidth}{!}{
		\begin{tabular}{|c|c|c|c|c|c|c|c|c|c|c|}
			\hline
			\textbf{} & 1SVM & k1SVM & SVDD & kSVDD & kNFST & 1vSet & 1MPM & DMPM & OC-ACNN & Proposed \\ \hline
			MOBIO     & \begin{tabular}[c]{@{}c@{}}0.632\\ (0.004)\end{tabular} & \begin{tabular}[c]{@{}c@{}}0.748\\ (0.004)\end{tabular} & \begin{tabular}[c]{@{}c@{}}0.582\\ (0.007)\end{tabular} & \begin{tabular}[c]{@{}c@{}}0.763\\ (0.013)\end{tabular} & \begin{tabular}[c]{@{}c@{}}0.560\\ (0.003)\end{tabular} & \begin{tabular}[c]{@{}c@{}}0.670\\ (0.005)\end{tabular} & \begin{tabular}[c]{@{}c@{}}0.768\\ (0.003)\end{tabular} & \begin{tabular}[c]{@{}c@{}}0.825\\ (0.007)\end{tabular} & \begin{tabular}[c]{@{}c@{}}0.938\\ (0.005)\end{tabular} & \textbf{\begin{tabular}[c]{@{}c@{}}0.998\\ (0.003)\end{tabular}} \\ \hline
			UMDAA-01  & \begin{tabular}[c]{@{}c@{}}0.622\\ (0.002)\end{tabular} & \begin{tabular}[c]{@{}c@{}}0.731\\ (0.009)\end{tabular} & \begin{tabular}[c]{@{}c@{}}0.615\\ (0.018)\end{tabular} & \begin{tabular}[c]{@{}c@{}}0.701\\ (0.009)\end{tabular} & \begin{tabular}[c]{@{}c@{}}0.567\\ (0.012)\end{tabular} & \begin{tabular}[c]{@{}c@{}}0.593\\ (0.017)\end{tabular} & \begin{tabular}[c]{@{}c@{}}0.816\\ (0.003)\end{tabular} & \begin{tabular}[c]{@{}c@{}}0.869\\ (0.001)\end{tabular} & \begin{tabular}[c]{@{}c@{}}0.891\\ (0.002)\end{tabular} & \textbf{\begin{tabular}[c]{@{}c@{}}0.954\\ (0.005)\end{tabular}} \\ \hline
			UMDAA-02  & \begin{tabular}[c]{@{}c@{}}0.614\\ (0.008)\end{tabular} & \begin{tabular}[c]{@{}c@{}}0.649\\ (0.004)\end{tabular} & \begin{tabular}[c]{@{}c@{}}0.515\\ (0.007)\end{tabular} & \begin{tabular}[c]{@{}c@{}}0.550\\ (0.007)\end{tabular} & \begin{tabular}[c]{@{}c@{}}0.556\\ (0.003)\end{tabular} & \begin{tabular}[c]{@{}c@{}}0.538\\ (0.003)\end{tabular} & \begin{tabular}[c]{@{}c@{}}0.722\\ (0.006)\end{tabular} & \begin{tabular}[c]{@{}c@{}}0.760\\ (0.007)\end{tabular} & \begin{tabular}[c]{@{}c@{}}0.735\\ (0.009)\end{tabular} & \textbf{\begin{tabular}[c]{@{}c@{}}0.813\\ (0.006)\end{tabular}} \\ \hline
		\end{tabular}
	}
\end{table*}

We consider the following methods from the active authentication literature for comparison:\\
\noindent \textbf{1. Linear OCSVM (1SVM):} One-class SVM (OC-SVM) as formulated in \cite{scholkopf2001estimating} is trained with a linear kernel on features of given user.\\
\noindent \textbf{2. Linear SVDD (SVDD):} Support vector data descriptor (SVDD) with a linear kernel as formulated in \cite{tax2004support} is trained on features of given user.\\
\noindent \textbf{3. Kernel OCSVM (k1SVM):} OC-SVM as formulated in \cite{scholkopf2001estimating} is trained on the given features with a radial basis function (RBF) kernel.\\
\noindent \textbf{4. Kernel SVDD (kSVDD):}  SVDD with RBF kernel as formulated in \cite{tax2004support} is trained on given features.\\
\noindent \textbf{5. One-class kNFST (kNFST):} Kernel null foley-sammon transform is used as proposed in \cite{bodesheim2013kernel}. kNFST finds a single null-space direction in feature space where intra-class distance of the class is low.\\
\noindent \textbf{6. One-vs-set Machines (1vSet):} As proposed in \cite{bendale2016towards}, two hyper-planes are optimized to enclose given category features within a slab in feature space.\\
\noindent \textbf{7. Single-MPM (1MPM):} This formulation as proposed in \cite{ghaoui2003robust}, considers second order statistics to learn a better hyperplane that separates origin from the one-class data in the feature space.\\
\noindent \textbf{8. Dual-MPM (DMPM):} Proposed in \cite{perera2018dual}, DMPM extends the 1MPM formulation by learning an additional hyperplane that better encloses given features.\\
\noindent \textbf{9. OC-ACNN:} Method proposed in \cite{oza2019active}, develop a deep convolutional neural network based one-class classifier by using Gaussian as pseudo-negative samples and regularizing the feature space with a decoder network.

Table.~\ref{table:ada} compares the performance of the proposed method with the state-of-the-art active authentication models. Out of all methods, 1SVM's and SVDD's performances are the lowest. Both of these methods are able to improve the performance when the kernel trick is incorporated into their formulations as shown by k1SVM and kSVDD, respectively. 1vSet and kNFST prove competitive against the classical one-class formulations such as OC-SVM and SVDD. Out of all the methods based on hyperplane optimization formulation, the MPM-based methods clearly outperform all the others. Specifically, DMPM is able to outperform 1MPM by $\sim$5\%, $\sim$6\% and $\sim$4\%, respectively on MOBIO, UMDAA-01 and UMDAA-02 datasets. OC-ACNN provides a considerable improvement compared to DMPM on MOBIO and UMDAA-01, but under performs on UMDAA-02. The proposed method outperforms all the other methods. More precisely, the proposed method observes $\sim$6\%, $\sim$6\% and $\sim$5\% improvement over the next best baseline on MOBIO, UMDAA-01 and UMDAA-02, respectively. This improvement can be largely attributed to the fact that federated learning framework enables privacy preserving collaboration among devices that results in a better active authentication system compared to the traditional one-class modeling based methods.

\subsection{Ablation analysis}\label{subsec:ablation_analysis}

\subsubsection{Impact of number of unknown}
Table.~\ref{table:no_unknown} shows the impact of varying the number of unknown/unauthorized users on the authentication system. For the experiment, we consider the UMDAA-01 dataset with all the implementation detail kept the same as described in Sec.~\ref{subsec:implementation_details} and the number of enrolled users are fixed to 25. As evident from the table, the performance decreases as we increase the number of unknown/unauthorized user during testing.

\subsection{Fedarated/split learning vs proposed method}\label{subsec:federated_average_vs_proposed_method}


\subsubsection{Performance}

\begin{figure}[b!]
	\centering
	\includegraphics[width=.85\linewidth]{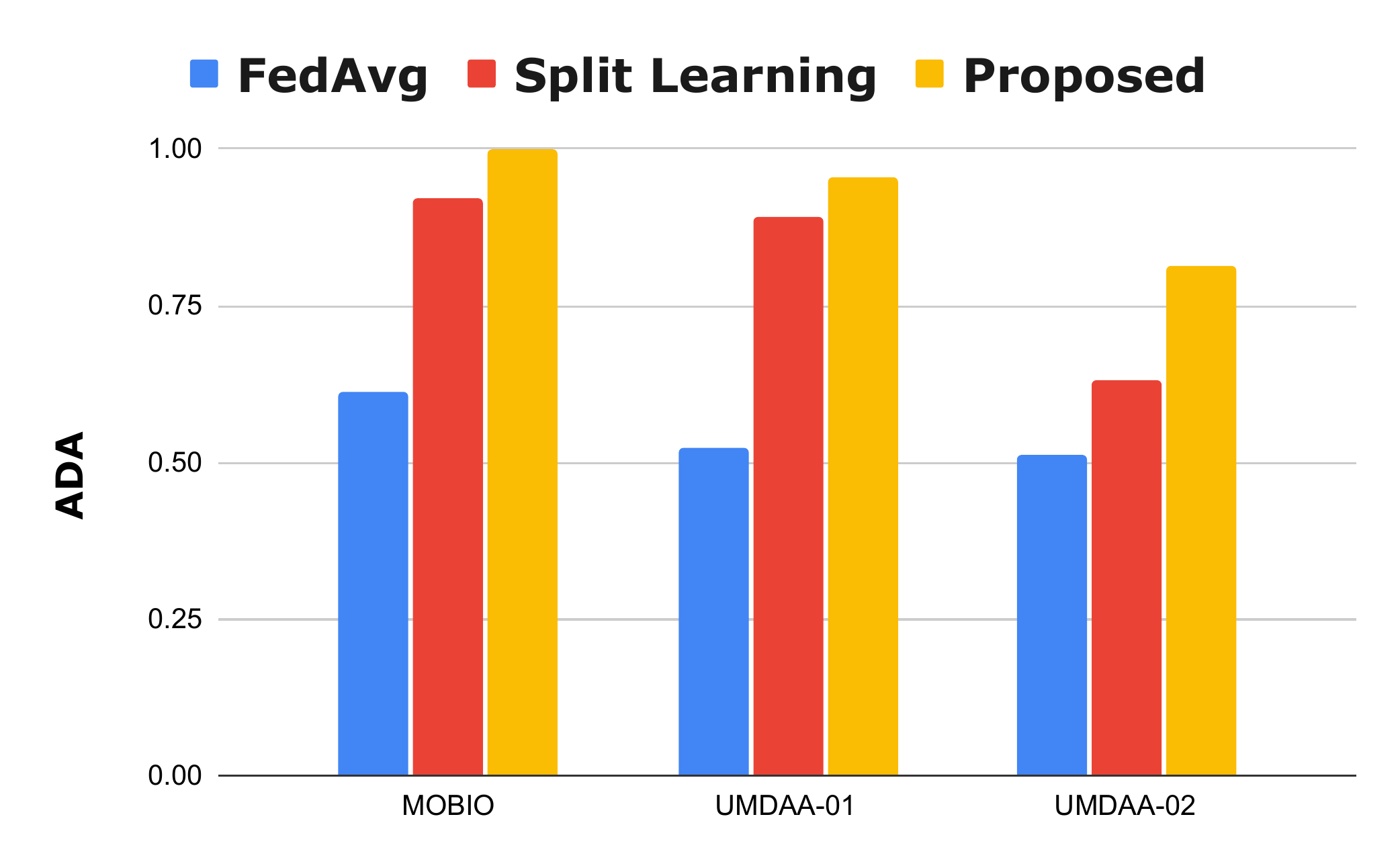}
	\caption{Comparing the performance between FedAvg, Split Learning \cite{gupta2018distributed} and the proposed method on the MOBIO, UMDAA-01 and UMDAA-02 datasets.}
	\label{fig:fedavg_vs_proposed}
\end{figure}

We compare the performance of FL and SL approaches with the proposed method. We the methods on all three datasets using the experimental protocol described in Sec.~\ref{subsec:datasets}. As can be seen from Fig.~\ref{fig:fedavg_vs_proposed}, the proposed method is able to perform much better compared to both FedAvg and Split Learning Approach (SLA) \cite{gupta2018distributed} on all three datasets. In the case of MOBIO, both FedAvg and SLA perform the best compared to the other two datasets, providing average detection accuracy of 61.2\% and 92\%, respectively. In comparison, the proposed approach is able to achieve 99.8\% average detection accuracy, resulting in nearly 38\% and 7\% improvement on the MOBIO dataset, respectively. For the slightly challenging UMDAA-01 dataset, when the authentication model is trained using FedAvg and SLA, the model achieves the performance of 52.4\% and 89\%, respectively. Compared to FedAvg and SLA, the proposed approach achieves 95.4\% average detection accuracy. Similarly, FedAvg and SLA perform about 51\% and 62\%, respectively on  the most challenging  UMDAA-02 dataset.  Whereas, the proposed approach achieves 81.2\% average detection accuracy, resulting in respective 29\% and 19\% improvement. As discussed in Sec.~\ref{subsec:challenges}, the major reason why FL/SL methods perform poorly is due to the highly non-IID nature (i.e. $qIID=0$) of the federated active authentication problem. Though SLA comes very close to the performance of the proposed method, it still requires multiple rounds of communication between device and server. In contrast, the proposed approach requires only one round of communication between device and server.

\begin{table}[t!]
	\centering
	\huge
	\caption{Impact on average detection accuracy with increasing the number of unknown/unauthorized users for the UMDAA-01 dataset.}
	\label{table:no_unknown}
	\resizebox{.70\linewidth}{!}{
		\begin{tabular}{|c|c|c|c|c|}
			\hline
			\begin{tabular}[c]{@{}c@{}}Number of \\ Unknown User\end{tabular} & 10 & 15 & 20 & 25 \\ \hline
			UMDAA-01                                                        & \begin{tabular}[c]{@{}c@{}}0.983\\ (0.003)\end{tabular} & \begin{tabular}[c]{@{}c@{}}0.976\\ (0.003)\end{tabular} & \begin{tabular}[c]{@{}c@{}}0.963\\ (0.002)\end{tabular} & \begin{tabular}[c]{@{}c@{}}0.954\\ (0.005)\end{tabular} \\ \hline
		\end{tabular}
	}
\end{table}

\section{Conclusion}\label{sec:conclusion}
We proposed a novel approach for user active authentication based on federated and split learning frameworks, called Federate Active Authentication. We point out the limitations of existing active authentication methods that model it as a one-class classification problem. The proposed method utilizes the federated/split learning framework to go beyond the one-class assumption for user active authentication. We also show that existing federated/split learning algorithms perform poorly on the federated active authentication setting. To address these issues, we proposed a novel method that extracts feature statistics of each user and trains a classification network to perform a multi-class classification, resulting in an efficient training strategy and improved authentication model. The proposed method is evaluated on three publicly available datasets and it is shown that it can perform better compared to both one-class modeling based active authentication methods and existing federated/split learning approaches. Furthermore, we analyze the effectiveness of the proposed method under varying number of enrolled and unknown/unauthorized users.

{\small
\bibliographystyle{ieee}
\bibliography{egbib}

\begin{thebibliography}{10}\itemsep=-1pt

\bibitem{bendale2016towards}
A.~Bendale and T.~E. Boult.
\newblock Towards open set deep networks.
\newblock In {\em Proceedings of the IEEE conference on computer vision and
  pattern recognition}, pages 1563--1572, 2016.

\bibitem{bodesheim2013kernel}
P.~Bodesheim, A.~Freytag, E.~Rodner, M.~Kemmler, and J.~Denzler.
\newblock Kernel null space methods for novelty detection.
\newblock In {\em Proceedings of the IEEE conference on computer vision and
  pattern recognition}, pages 3374--3381, 2013.

\bibitem{dey2016extreme}
D.~K. Dey and J.~Yan.
\newblock {\em Extreme value modeling and risk analysis: methods and
  applications}.
\newblock CRC Press, 2016.

\bibitem{fathy2015face}
M.~E. Fathy, V.~M. Patel, and R.~Chellappa.
\newblock Face-based active authentication on mobile devices.
\newblock In {\em 2015 IEEE International Conference on Acoustics, Speech and
  Signal Processing (ICASSP)}, pages 1687--1691. IEEE, 2015.

\bibitem{frank2012touchalytics}
M.~Frank, R.~Biedert, E.~Ma, I.~Martinovic, and D.~Song.
\newblock Touchalytics: On the applicability of touchscreen input as a
  behavioral biometric for continuous authentication.
\newblock {\em IEEE transactions on information forensics and security},
  8(1):136--148, 2012.

\bibitem{ghaoui2003robust}
L.~E. Ghaoui, M.~I. Jordan, and G.~R. Lanckriet.
\newblock Robust novelty detection with single-class mpm.
\newblock In {\em Advances in neural information processing systems}, pages
  929--936, 2003.

\bibitem{gunther20132013}
M.~G{\"u}nther, A.~Costa-Pazo, C.~Ding, E.~Boutellaa, G.~Chiachia, H.~Zhang,
  M.~de~Assis~Angeloni, V.~{\v{S}}truc, E.~Khoury, E.~Vazquez-Fernandez, et~al.
\newblock The 2013 face recognition evaluation in mobile environment.
\newblock In {\em 2013 International Conference on Biometrics (ICB)}, pages
  1--7. IEEE, 2013.

\bibitem{gupta2018distributed}
O.~Gupta and R.~Raskar.
\newblock Distributed learning of deep neural network over multiple agents.
\newblock {\em Journal of Network and Computer Applications}, 116:1--8, 2018.

\bibitem{kumar2016continuous}
R.~Kumar, V.~V. Phoha, and A.~Serwadda.
\newblock Continuous authentication of smartphone users by fusing typing,
  swiping, and phone movement patterns.
\newblock In {\em 2016 IEEE 8th International Conference on Biometrics Theory,
  Applications and Systems (BTAS)}, pages 1--8. IEEE, 2016.

\bibitem{lanckriet2002minimax}
G.~Lanckriet, L.~E. Ghaoui, C.~Bhattacharyya, and M.~I. Jordan.
\newblock Minimax probability machine.
\newblock In {\em Advances in neural information processing systems}, pages
  801--807, 2002.

\bibitem{mahbub2016active}
U.~Mahbub, S.~Sarkar, V.~M. Patel, and R.~Chellappa.
\newblock Active user authentication for smartphones: A challenge data set and
  benchmark results.
\newblock In {\em 2016 IEEE 8th International Conference on Biometrics Theory,
  Applications and Systems (BTAS)}, pages 1--8. IEEE, 2016.

\bibitem{mcmahan2016communication}
H.~B. McMahan, E.~Moore, D.~Ramage, S.~Hampson, et~al.
\newblock Communication-efficient learning of deep networks from decentralized
  data.
\newblock {\em arXiv preprint arXiv:1602.05629}, 2016.

\bibitem{mohri2019agnostic}
M.~Mohri, G.~Sivek, and A.~T. Suresh.
\newblock Agnostic federated learning.
\newblock {\em arXiv preprint arXiv:1902.00146}, 2019.

\bibitem{oza2019active}
P.~Oza and V.~M. Patel.
\newblock Active authentication using an autoencoder regularized cnn-based
  one-class classifier.
\newblock In {\em 2019 14th IEEE International Conference on Automatic Face \&
  Gesture Recognition (FG 2019)}, pages 1--8. IEEE, 2019.

\bibitem{Parkhi15}
O.~M. Parkhi, A.~Vedaldi, and A.~Zisserman.
\newblock Deep face recognition.
\newblock In {\em British Machine Vision Conference}, 2015.

\bibitem{patel2016continuous}
V.~M. Patel, R.~Chellappa, D.~Chandra, and B.~Barbello.
\newblock Continuous user authentication on mobile devices: Recent progress and
  remaining challenges.
\newblock {\em IEEE Signal Processing Magazine}, 33(4):49--61, 2016.

\bibitem{perera2017extreme}
P.~Perera and V.~M. Patel.
\newblock Extreme value analysis for mobile active user authentication.
\newblock In {\em 2017 12th IEEE International Conference on Automatic Face \&
  Gesture Recognition (FG 2017)}, pages 346--353. IEEE, 2017.

\bibitem{perera2018dual}
P.~Perera and V.~M. Patel.
\newblock Dual-minimax probability machines for one-class mobile active
  authentication.
\newblock In {\em 2018 IEEE 9th International Conference on Biometrics Theory,
  Applications and Systems (BTAS)}, pages 1--8. IEEE, 2018.

\bibitem{poirot2019split}
M.~G. Poirot, P.~Vepakomma, K.~Chang, J.~Kalpathy-Cramer, R.~Gupta, and
  R.~Raskar.
\newblock Split learning for collaborative deep learning in healthcare.
\newblock {\em arXiv preprint arXiv:1912.12115}, 2019.

\bibitem{scholkopf2001estimating}
B.~Sch{\"o}lkopf, J.~C. Platt, J.~Shawe-Taylor, A.~J. Smola, and R.~C.
  Williamson.
\newblock Estimating the support of a high-dimensional distribution.
\newblock {\em Neural computation}, 13(7):1443--1471, 2001.

\bibitem{seddik2020random}
M.~E.~A. Seddik, C.~Louart, M.~Tamaazousti, and R.~Couillet.
\newblock Random matrix theory proves that deep learning representations of
  gan-data behave as gaussian mixtures.
\newblock {\em arXiv preprint arXiv:2001.08370}, 2020.

\bibitem{serwadda2013verifiers}
A.~Serwadda, V.~V. Phoha, and Z.~Wang.
\newblock Which verifiers work?: A benchmark evaluation of touch-based
  authentication algorithms.
\newblock In {\em 2013 IEEE Sixth International Conference on Biometrics:
  Theory, Applications and Systems (BTAS)}, pages 1--8. IEEE, 2013.

\bibitem{simonyan2014very}
K.~Simonyan and A.~Zisserman.
\newblock Very deep convolutional networks for large-scale image recognition.
\newblock {\em arXiv preprint arXiv:1409.1556}, 2014.

\bibitem{tax2004support}
D.~M. Tax and R.~P. Duin.
\newblock Support vector data description.
\newblock {\em Machine learning}, 54(1):45--66, 2004.

\bibitem{thapa2020splitfed}
C.~Thapa, M.~A.~P. Chamikara, and S.~Camtepe.
\newblock Splitfed: When federated learning meets split learning.
\newblock {\em arXiv preprint arXiv:2004.12088}, 2020.

\bibitem{vepakomma2018split}
P.~Vepakomma, O.~Gupta, T.~Swedish, and R.~Raskar.
\newblock Split learning for health: Distributed deep learning without sharing
  raw patient data.
\newblock {\em arXiv preprint arXiv:1812.00564}, 2018.

\end{thebibliography}
}

\end{document}